\documentclass[sigconf]{acmart}
\AtBeginDocument{%
  }

\setcopyright{acmlicensed}
\copyrightyear{2025}
\acmYear{2025}

\newcommand{\zj}[1]{\textcolor{black}{#1}}
\acmSubmissionID{226}

\usepackage{algorithm}      
\usepackage{algpseudocode} 
\usepackage{algorithmicx} 
\usepackage{multirow} 
\usepackage{tabularx}
\usepackage{makecell}

\usepackage{multirow}
\usepackage{tabularx}
\usepackage{booktabs}
\usepackage{graphicx}
\usepackage{caption} 
\begin{document}

\title{OoDDINO:A Multi-level Framework for Anomaly Segmentation on Complex Road Scenes }

\author{Yuxing Liu}
\affiliation{%
  \institution{College of Computer Science and Artificial Intelligence, Southwest Minzu University}
  \city{Chengdu}
  \state{Sichuan}
  \country{China}}
\email{lyxlyx_47@outlook.com}

\author{Ji Zhang}
\affiliation{%
  \institution{College of Computer Science and Artificial Intelligence, Southwest Minzu University}
  \city{Chengdu}
  \state{Sichuan}
  \country{China}}
\affiliation{%
  \institution{Engineering Research Center of Sustainable Urban Intelligent Transportation, Ministry of Education, China}
  \country{}}
\email{jizhang901@gmail.com}

\author{Xuchuan Zhou}
\affiliation{%
  \institution{College of Computer Science and Artificial Intelligence, Southwest Minzu University}
  \city{Chengdu}
  \state{Sichuan}
  \country{China}}
\email{xczhou@swun.edu.cn}

\author{Jingzhong Xiao}
\authornotemark[1]
\affiliation{%
  \institution{College of Computer Science and Artificial Intelligence, Southwest Minzu University}
  \city{Chengdu}
  \state{Sichuan}
  \country{China}}
\email{21700013@swun.edu.cn}

\author{Huimin Yang}
\affiliation{%
  \institution{College of Computer Science and Artificial Intelligence, Southwest Minzu University}
  \city{Chengdu}
  \state{Sichuan}
  \country{China}}
\email{yhm653750@gmail.com}

\author{Jiaxin Zhong}
\affiliation{%
  \institution{College of Computer Science and Artificial Intelligence, Southwest Minzu University}
  \city{Chengdu}
  \state{Sichuan}
  \country{China}}
\email{zjxzjx5611@outlook.com}

\thanks{*Corresponding author: Jingzhong Xiao}

\renewcommand{\shortauthors}{Trovato et al.}

\begin{abstract}
 Anomaly segmentation aims to identify Out-of-Distribution (OoD) anomalous objects within images. Existing pixel-wise methods typically assign anomaly scores individually and employ a global thresholding strategy to segment anomalies. Despite their effectiveness, these approaches encounter significant challenges in real-world applications: (1) neglecting spatial correlations among pixels within the same object, resulting in fragmented segmentation; (2) variability in anomaly score distributions across image regions, causing global thresholds to either generate false positives in background areas or miss segments of anomalous objects. In this work, we introduce OoDDINO, a novel multi-level anomaly segmentation framework designed to address these limitations through a coarse-to-fine anomaly detection strategy. OoDDINO combines an uncertainty-guided anomaly detection model with a pixel-level segmentation model within a two-stage cascade architecture. Initially, we propose an Orthogonal Uncertainty-Aware Fusion Strategy (OUAFS) that sequentially integrates multiple uncertainty metrics with visual representations, employing orthogonal constraints to strengthen the detection model's capacity for localizing anomalous regions accurately. Subsequently, we develop an Adaptive Dual-Threshold Network (ADT-Net), which dynamically generates region-specific thresholds based on object-level detection outputs and pixel-wise anomaly scores. This approach allows for distinct thresholding strategies within foreground and background areas, achieving fine-grained anomaly segmentation. The proposed framework is compatible with other pixel-wise anomaly detection models, which act as a plug-in to boost the performance. Extensive experiments on two benchmark datasets validate our framework's superiority and compatibility over state-of-the-art methods. Source code is available at: \url{https://github.com/OoDDINO/OoD-DINO}.
\end{abstract}



\setcopyright{acmcopyright}
\copyrightyear{2025}
\acmYear{2025}
\acmDOI{XXXXXXX.XXXXXXX}

\acmConference[ACM MM'25]{ACM Multimedia conference}{October 27--31, 2025}{Dublin, Ireland}
\acmBooktitle{ACM MM'25: ACM Multimedia conference, October 27--31, 2025, ACM MM, Dublin, Ireland}

\maketitle
\section{Introduction}
Semantic segmentation, as a foundational task in computer vision, aims at classifying each pixel into predefined visual categories \cite{cheng2022gsrformer,long2015fully}. Despite remarkable advances, existing segmentation methods are primarily restricted to recognizing objects within pre-established training distributions, limiting their applicability to open-set environments. In real-world, open-set contexts, particularly safety-critical domains such as autonomous driving \cite{qiao2022real,qiao2022swnet}, segmentation models inevitably encounter out-of-distribution (OoD) or anomalous objects not represented in training sets. The diverse and unpredictable nature of these anomalous objects creates significant challenges, making it impractical to construct exhaustive datasets \cite{he2023damo,hu2021region,lan2023procontext,li2023longshortnet}. Therefore, anomaly segmentation has emerged as a critical extension of semantic segmentation, aimed explicitly at identifying and localizing OoD objects through pixel-level detection \cite{lis2019detecting,caesar2020nuscenes,cordts2016cityscapes,hein2019relu,nguyen2015deep,yu2020bdd100k,liu2023residual}.
\begin{figure}[t]
  \centering
  \includegraphics[width=0.47\textwidth]{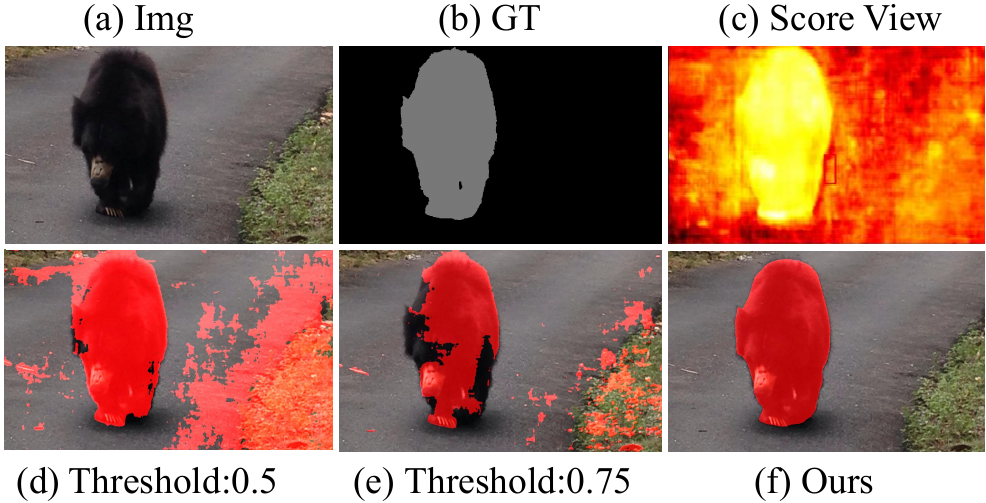}
  \caption{(a) Input image. (b)Ground Truth. (c)The anomaly score heatmap reveals varying levels of abnormality across different regions. (d) Low threshold (50\%) detection, complete anomaly but high false positives. (e) High threshold (75\%) detection, reduced noise but incomplete anomaly. (f) Region-adaptive segmentation result.}
  \label{fig:motivation}
  \vspace{-4mm}
\end{figure}

Existing anomaly segmentation methods generally adopt pixel-wise anomaly scoring followed by a global thresholding mechanism for segmentation decisions. Although straightforward, this methodology faces two critical limitations in practice. First, by processing pixels independently and disregarding inherent spatial coherence within object regions, existing methods frequently generate fragmented segmentation outcomes \cite{rai2024mask2anomaly,zhao2024segment}. Second, due to substantial regional variability in anomaly score distributions, a single global threshold cannot simultaneously minimize false positives in background regions and avoid incomplete detection of anomalous objects \cite{zhao2024segment} (as shown in Fig.~\ref{fig:motivation}c). Specifically, lower thresholds effectively capture anomalies yet induce excessive false positives, whereas higher thresholds suppress noise but compromise anomaly completeness (as shown in Fig.~\ref{fig:motivation}d,e).

To overcome these challenges, anomaly segmentation methods should explicitly leverage object-level spatial priors, enabling models to focus selectively on relevant anomalous regions while effectively suppressing irrelevant background information \cite{rai2024mask2anomaly}. Recently, open-set object detection techniques, capable of simultaneously detecting known and unknown categories without explicit annotations for anomalies, have attracted considerable attention \cite{joseph2021towards,liu2024grounding,carion2020end,gu2021open,yao2022detclip,zareian2021open}. These methods provide valuable preliminary region proposals that preserve object coherence and substantially mitigate background interference, which motivates our research.
\begin{figure*}[]
    \centering
    \includegraphics[width=1\textwidth]{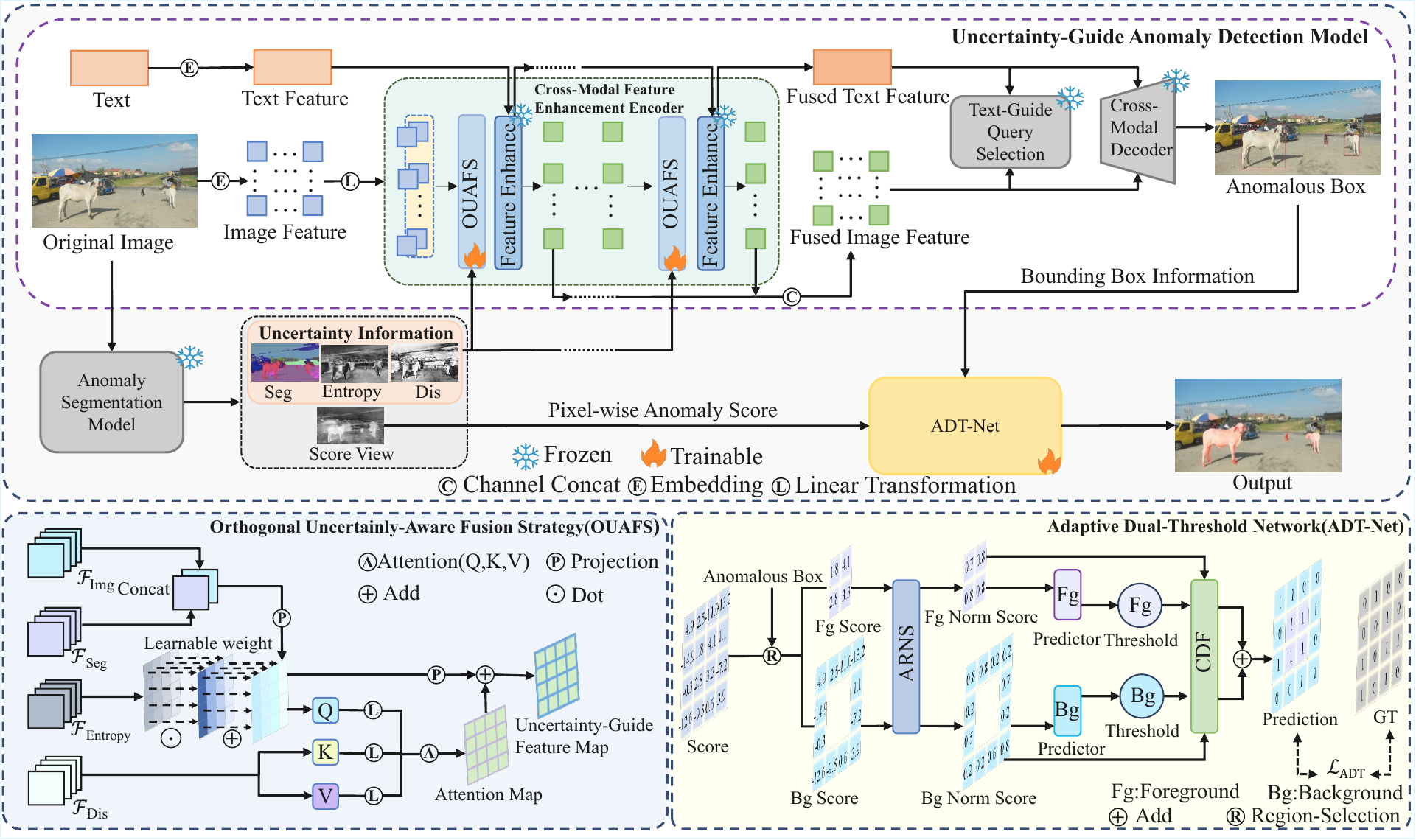} 
    \vspace{-5mm}
    \caption{\textbf{The OoDDINO framework integrates two complementary modules: Orthogonal Uncertainty-Aware Fusion Strategy (OUAFS) and Adaptive Dual-Threshold Network (ADT-Net). OUAFS enhances detection by sequentially fusing multi-dimensional uncertainty features, while ADT-Net dynamically generates region-specific thresholds to optimize pixel-level anomaly selection.}}
    \label{fig:framework}
    \vspace{-2mm}
\end{figure*}

In this paper, we propose OoDDINO, a novel multi-level anomaly segmentation framework specifically tailored for open-world road scenarios. Specifically, our framework integrates uncertainty-guided object-level anomaly detection with adaptive refinement at the pixel level, facilitating a hierarchical, coarse-to-fine segmentation paradigm. In the first stage, we introduce the Orthogonal Uncertainty-Aware Fusion Strategy (OUAFS), designed to sequentially fuse multiple uncertainty-driven features with visual representations under orthogonal constraints, thereby enhancing object-level anomaly detection accuracy. Subsequently, to overcome limitations inherent to global thresholding, we propose the Adaptive Dual-Threshold Network (ADT-Net), which dynamically generates region-specific thresholds by jointly leveraging object-level detection outputs and pixel-wise anomaly scores, thus enabling precise pixel-level anomaly classification. Furthermore, the proposed framework is highly modular and can seamlessly integrate with existing anomaly segmentation methods, significantly improving their performance. Our contributions are summarized as follows:
\begin{itemize}
    \item To incorporate object-level spatial priors into anomaly segmentation methods, we propose a novel multi-level anomaly segmentation framework OoD-DINO, which adaptively integrates uncertainty-guided anomaly detection models with anomaly segmentation approaches, establishing a coarse-to-fine precise anomaly segmentation paradigm.
    
    \item We introduce the Orthogonal Uncertainty-Aware Fusion Strategy (OUAFS), a novel feature fusion method leveraging multi-dimensional uncertainty information to enhance the precision of anomaly region detection. Specifically, OUAFS employs sequential multi-stage fusion guided by orthogonal constraints to maximize complementary feature integration while reducing redundancy.
    
    \item To overcome inherent limitations of global thresholding, we propose the Adaptive Dual-Threshold Network (ADT-Net), a novel thresholding mechanism that dynamically generates differentiated, region-specific thresholds for foreground and background regions by integrating object-level detections and pixel-wise anomaly scores, thus significantly improving anomaly segmentation granularity.
    
    \item Comprehensive experiments on the SMIYC and RoadAnomaly datasets demonstrate that our proposed framework, based on different baseline methods, consistently achieves state-of-the-art performance. 
\end{itemize}
\section{Related Work}
\subsection{Anomaly Segmentation}
Existing anomaly segmentation approaches\cite{kendall2017uncertainties,lee2018simple,hendrycks2016baseline,GalGhahramani2016Dropout,LakshminarayananPritzelBlundell2017Simple}, can be categorized into two classes: uncertainty-based and energy-based.

Uncertainty-based methods identify anomalous regions by predicting areas with high uncertainty \cite{kendall2017uncertainties,lee2018simple,hendrycks2016baseline}. The early approach estimates uncertainty using predicted softmax values, resulting in remarkable performance in image-level tasks \cite{HendrycksGimpel2017Baseline}. However, this method often struggles with accurately handling the boundary pixels of anomalous objects, leading to degraded anomaly detection performance\cite{Tian2022pixelwise}. To address this limitation, \cite{DiBiase2021pixelwise} leverages predictions from multiple models to estimate uncertainty, while MC Dropout \cite{GalGhahramani2016Dropout} utilizes the randomness of dropout layers to obtain uncertainty estimates. Although these methods have somewhat improved the performance of anomaly detection, they achieve lower accuracy in anomaly segmentation tasks\cite{LakshminarayananPritzelBlundell2017Simple}.

Energy-based methods employ energy functions to evaluate the anomaly level of individual pixels by assigning confidence scores. PEBAL \cite{Tian2022pixelwise} learns energy-based penalties through adversarial training with outlier exposure, RPL \cite{liu2023residual} introduces residual pattern learning with context-robust contrastive constraints. Despite their effectiveness, these methods face critical limitations\cite{nayal2023rba}. They heavily rely on the accuracy of in-distribution data modeling, essentially detecting anomalies by identifying deviations from known patterns rather than recognizing anomalous characteristics directly. 

Despite technical differences, both paradigms operate through pixel-wise inference with global thresholding for anomaly classification. They ignore semantic consistency and rely on naive thresholds, leading to fragmented segmentation and a trade-off between false positives and missed anomalies.

\subsection{Open-Set Object Detection}
Open-set object detection aims to identify objects belonging to known categories while detecting objects of unknown categories. GLIP \cite{GaoGengZhangMaFangZhangLiQiao2021Clip} treats object detection as a core problem and enhances semantic alignment learning by using additional underlying data. It expands the detector's vocabulary by introducing extensive text descriptions, thereby improving the recognition of unknown categories. GroundingDINO \cite{liu2024grounding} proposes an innovative cross-modal fusion mechanism that effectively bridges visual and textual representations through tightly-coupled feature interactions.

Open-set object detection has some ability to handle unknown categories, but the diversity and quality of the training data still limits it. The model's ability to generalize in real-world scenarios is restricted by too few object categories in the training set or insufficient scene variation. It becomes very challenging to effectively locate regions that contain Out-of-Distribution (OoD) objects \cite{miller2021uncertainty}.

In anomaly segmentation tasks,  the model aims to identify anomalous objects. Incorporating multimodal information (such as uncertainty and textual information) enhances the capability to detect unknown categories in an open-set detection, thereby mitigating background noise. This approach presents a potential solution. To the best of our knowledge, this is the first time an open-set detection model has been applied to anomaly segmentation.
\section{Methodology \label{sec:method}}
\subsection{Overall Framework}
In this paper, a multi-level anomaly segmentation framework, OoDDINO, is proposed to address the critical limitations in existing anomaly segmentation methods. By integrating object-level detection with pixel-level OoD identification, OoDDINO establishes a coarse-to-fine anomaly segmentation paradigm that progressively refines anomaly classification. As illustrated in Fig.~\ref{fig:framework}, the framework is constructed upon the open-set object detection model GroundingDINO\cite{liu2024grounding}, which works in parallel with an anomaly segmentation models to generate instance-level bounding boxes and pixel-level anomaly score maps, respectively. To enhance GroundingDINO's capability in accurately localizing anomalous objects, an Orthogonal Uncertainty-Aware Fusion Strategy (OUAFS) is integrated into multiple feature extraction stages, enriching anomaly representation by sequentially integrating various uncertainty features with visual representations under orthogonal constraints. For fine-grained anomaly pixel classification, \zj{an Adaptive Dual-Threshold Network (ADT-Net) is proposed to dynamically leverage object bounding boxes and anomaly score maps for generating region-specific thresholds, enabling high-sensitivity detection in anomalous regions while suppressing background noise.}

Specifically, our baseline anomaly detection architecture comprises three core components: (1) a Cross-Modal Feature Enhancement Encoder that extracts and fuses image features with uncertainty measures and textual embeddings to emphasize anomalous regions. (2) a Text-Guided Query Selection mechanism that prioritizes image-text relevant regions for improved detection precision. (3) a Cross-Modal Decoder that aligns visual and textual representations through attention mechanisms to generate object bounding boxes and classification labels. 
\subsection{Orthogonal Uncertainty-Aware Fusion Strategy\label{sec:OUAFS}}
\zj{Due to the insufficient diversity of objects and scenes in training datasets, existing open-set detection methods suffer from limited generalization ability in real-world scenarios, making it challenging to adapt to dynamic and uncertain anomaly detection settings}\cite{miller2021uncertainty}. To address this challenge, CF-MAD \cite{wang2022multimodal} detects OoD objects by integrating multi-modal information from various sources. While this approach yields promising results, it lacks precise control over the fusion process, leading to information overload or an imbalance in the contributions of different modalities.

In this paper, an Orthogonal Uncertainty-Aware Fusion Strategy (OUAFS) is proposed, which systematically integrates various uncertainty information with visual features and employs an orthogonal loss to optimize the fusion process across different modalities. Considering the differences between various uncertainty maps, \zj{the fusion method and order of the uncertainty maps are carefully designed to fully leverage the complementary information between modalities, as shown in Algorithm \ref{alg:ouafs}}, thereby enhancing anomaly feature representation. Specifically, three uncertainty maps are adopted: (1) Semantic segmentation map ${S_i}$ separates image pixels according to their semantic categories and provides explicit spatial information. To preserve semantic information, ${S_i}$  is directly concatenated with the image features. (2) Softmax entropy map ${E_i}$ reflects the uncertainty of each pixel and serves as a global-scale supplement. To avoid information redundancy, learnable weighted fusion is employed to dynamically adjust its contribution. (3) Softmax distance map \cite{DiBiase2021pixelwise} ${D_i}$ estimates the confidence of each pixel based on its distance from a reference distribution, capturing implicit features of the anomaly distribution. A cross-attention mechanism handles the complex nonlinear dependencies between this map and the semantic features of the image. After the fusion layer, a multi-head attention mechanism ensures spatial-level interaction between different features.


OUAFS evaluates the performance gain of each uncertainty map for anomaly detection by independently training models for each corresponding uncertainty map. Based on model performance, OUAFS serially fuses different uncertainty maps from low to high, ensuring the efficient utilization of complementary information between them. The performance gains of varying uncertainty maps are discussed in Section \ref{sec:abl}.

Features extracted from different uncertainty maps may contain redundant information during the fusion process. To enhance the diversity and complementarity of information between uncertainty maps, an orthogonal loss is proposed to encourage orthogonality between different uncertainty features, which can be formulated as follows:
\begin{equation}
\mathcal{L}_{\text{OUAFS}} = \lambda_1 \sum_{i=1}^{N-1} \left| \mathbf{F}_i \cdot \mathbf{F}_{i+1} \right| + \lambda_2 \sum_{i=1}^{N-1} \left( \mathbf{F}_i \cdot \mathbf{F}_{i+1} \right)^2
\end{equation}
where \( \lambda_1 \) and \( \lambda_2 \) are hyperparameters that control the strength of the two losses regularization terms, the \( N \) is the total number of modalities, and \( \mathbf{F}_i \) denotes the feature of the \( i \)-th modality.

\begin{algorithm}[h!]
  \caption{Orthogonal Uncertainty-Aware Fusion Strategy (OUAFS)}
  \label{alg:ouafs}
  \begin{algorithmic}[1]
  \Require 
  \Statex Image features: $\{\mathbf{F}_i\}_{i=1}^L$ where $\mathbf{F}_i \in \mathbb{R}^{H \times W \times C}$
  \Statex Uncertainty maps: $\{S_i, E_i, D_i\}_{i=1}^L$ (segmentation, entropy, distance)
  \Statex Text embeddings: $\mathbf{T} \in \mathbb{R}^{N \times D}$ \Comment{Embedded textual prompts}
  \Ensure Enhanced features: $\{\mathbf{\tilde{F}}_i\}_{i=1}^L$
  
  \State Initialize $\mathcal{E} \gets \emptyset$ \Comment{Enhanced feature list}
  \For{layer $i = 1$ to $L$}
      \State $\mathbf{X}_i \gets \text{Concat}(\mathbf{F}_i, S_i)$ 
      \State $\mathbf{X}_i \gets \mathbf{X}_i \odot \sigma(E_i)$ 
      \State $\mathbf{X}_i \gets \text{CrossAttn}(\mathbf{X}_i, D_i)$ 
      \State $\mathbf{\tilde{F}}_i \gets \text{CrossModalEncoder}(\mathbf{X}_i, \mathbf{T})$    \Comment{GroundingDINO-based encoding}
      \If{$i > 1$}
          \State $\mathbf{\tilde{F}}_i \gets \text{OrthoFuse}(\mathbf{\tilde{F}}_i, \mathcal{E}[i-1])$ \Comment{Orthogonal fusion}
      \EndIf
      
      \State $\mathcal{E}.\text{append}(\mathbf{\tilde{F}}_i)$
  \EndFor
  \State \Return $\mathcal{E}$
  \end{algorithmic}
  \end{algorithm}
\subsection{Adaptive Dual-Threshold Network\label{sec:ADTGN}}
\zj{Most existing anomaly segmentation methods treat every pixel in an image equally, classifying pixels with anomaly scores above a unified threshold as anomalous and those below normal. However, the likelihood of pixels being identified as anomalies varies across regions. Pixels within anomalous object regions are likelier to be anomalous, while those in background regions tend to be normal. Moreover, variations in objects and scenes across different images make it difficult for a fixed threshold to classify anomalous pixels accurately.} To address this, we propose ADT-Net, an adaptive dual-threshold network that integrates object detection and pixel-level anomaly scores to dynamically generate region-specific thresholds. By applying distinct thresholds inside and outside detected regions, ADT-Net achieves fine-grained anomaly selection.

Different anomaly segmentation models produce prediction scores with inconsistent distributions and ranges. To enhance ADT-Net's compatibility with diverse models, we propose an Adaptive-Region Normalization Strategy (ARNS)  that standardizes score distributions while preserving their discriminative properties between normal and anomalous regions.

Given an anomaly score map $I \in \mathbb{R}^{H \times W}$, we define the foreground mask as $\mathcal{M}_{\text{fg}} = \mathbf{1} \in \{1\}^{H \times W}$ and the background mask as \(\mathcal{M}_{\text{bg}} = 1 - \mathcal{M}_{\text{fg}}\), which is derived from detection proposals.
ADT-Net normalizes $I$ through piecewise nonlinear transformation to obtain $I_{\text{norm}}$:
\begin{equation}
I_{\text{norm}}(i,j) = 
\begin{cases} 
\frac{0.5}{1 + e^{-(I(i,j)-\mu_{\text{fg})}}} + \alpha, & \text{if } \mathcal{M}_{\text{fg}}(i,j)=1 \\
\frac{0.5}{1 + e^{-(I(i,j)-\mu_{\text{bg})}}} + \alpha, & \text{otherwise}
\end{cases}
\label{eq:eq}
\end{equation}
where $I_{\text{norm}}(i,j)$ are normalized to the range [$\alpha$, $\alpha$ + 0.5] to enhance the stability of the training process. The value of $\alpha$ is discussed in \ref{sec:sec3} section. $\mu_{\text{fg}}$ and $\mu_{\text{bg}}$ are computed as the mean anomaly scores within foreground and background regions, respectively:
\[
\mu_c = \frac{\sum_{i,j} I(i,j) \cdot \mathcal{M}_c(i,j)}{\sum_{i,j} \mathcal{M}_c(i,j)}, \quad c \in \{\text{fg}, \text{bg}\}.
\]

In ADT-Net, the architecturally identical foreground predictor $\mathcal{F}_{\theta}$ and background predictor $\mathcal{F}_{\phi}$ are utilized to process the normalized anomaly maps $I_{\text{norm}}$ as input, generating foreground threshold $T_{\text{fg}}$ and background threshold $T_{\text{bg}}$ respectively:
\begin{equation}
T_{\text{fg}} = \mathcal{F}_{\theta}(I_{\text{norm}} \odot \mathcal{M}_{\text{fg}})
\end{equation}
\begin{equation}
T_{\text{bg}} = \mathcal{F}_{\phi}(I_{\text{norm}} \odot (1 - \mathcal{M}_{\text{fg}}))
\end{equation}

Based on thresholds $T_{\text{fg}}$ and $T_{\text{bg}}$, pixels from different regions can be precisely classified, where those exceeding the thresholds are identified as anomalies while others are considered normal. Considering that this hard classification process is non-differentiable, we adopt a linear approximation to relax the binarization operation, enabling the gradient of thresholds to be backpropagated. Specifically, a Cumulative Distribution Function (CDF) is employed to model the anomaly scores, which can be formulated as follows:
{\small
\begin{equation}
P(y=1|I_{\text{norm}},T) = 
\begin{cases} 
\frac{I_{\text{norm}} - T_{\text{fg}} + \delta}{\delta}, & \mathcal{M}_{\text{fg}}=1,~T_{\text{fg}} \leq I_{\text{norm}} < T_{\text{fg}}+\delta \\[1ex]
\frac{I_{\text{norm}} - T_{\text{bg}} + \delta}{\delta}, & \mathcal{M}_{\text{fg}}=0,~T_{\text{bg}}-\delta < I_{\text{norm}} \leq T_{\text{bg}} \\[1ex]
1, & I_{\text{norm}} \geq T_{\text{fg}}+\delta ~\text{or}~ I_{\text{norm}} > T_{\text{bg}}+\delta \\
0, & \text{otherwise}
\end{cases}
\end{equation}
}
where $\delta$ controls the transition window width, empirically set to 0.1. This formulation allows for smooth gradient propagation while approximating a hard thresholding operation during inference. The final anomaly segmentation map is obtained by applying a threshold of 0.5 to $P(y=1|I_{\text{norm}},T)$ during inference.

Finally, a compound loss is proposed to jointly optimize both thresholds:
\begin{equation}
\mathcal{L}_{\text{ADT-Net}} =  \mathcal{L}_{\text{CE}}(P_{\text{fg}}, y) +  \mathcal{L}_{\text{CE}}(P_{\text{bg}}, 1-y) + \gamma \|T_{\text{fg}} - T_{\text{bg}}\|_2
\end{equation}
where $\mathcal{L}_{\text{CE}}$ denotes cross-entropy loss. The third term enforces threshold divergence to prevent decision boundary overlap. 
\subsection{Loss Function}
The proposed framework is optimized through a multi-task learning objective that integrates detection accuracy \cite{liu2024grounding}, feature fusion quality, and adaptive thresholding performance. Our composite loss function comprises three essential components:
\begin{equation}
  \mathcal{L}_{\text{Total}} = \lambda_{\text{detect}} \mathcal{L}_{\text{detect}} + \lambda_{\text{orth}} \mathcal{L}_{\text{orth}} + \lambda_{\text{ADT}} \mathcal{L}_{\text{ADT}}
  \end{equation}
where $\lambda_{\text{detect}}$, $\lambda_{\text{orth}}$, and $\lambda_{\text{ADT}}$ are parameters to balance the loss weights. The detection loss $\mathcal{L}_{\text{detect}}$ preserves the fundamental grounding capability, while $\mathcal{L}_{\text{orth}}$ represents the orthogonal fusion loss (Sec.~\ref{sec:OUAFS}). The term $\mathcal{L}_{\text{ADT}}$ corresponds to the ADT-Net loss (Sec.~\ref{sec:ADTGN}). 
\begin{table*}[t]
  \centering
  \caption{\textbf{Comparison with state-of-the-art methods on SMIYC benchmark. Best results in \textbf{bold}, second-best \underline{underlined}.} }
  \label{tab:smlyc_results}
  \resizebox{\textwidth}{!}{
  \begin{tabular}{l|c|ccccc|ccccc}
  \toprule
  \textbf{Method} & \textbf{Venue} & \multicolumn{5}{c|}{\textbf{AnomalyTrack}} & \multicolumn{5}{c}{\textbf{ObstacleTrack}} \\
  \cmidrule(lr){3-7} \cmidrule(lr){8-12}
   & & AP ↑ & FPR ↓ & sIoU ↑ & PPV ↑ & F1 ↑ & AP ↑ & FPR ↓ & sIoU ↑ & PPV ↑ & F1 ↑ \\
  \midrule
  Emb. Density\cite{blum2021fishyscapes} & IJCV'21 & 37.5 & 70.8 & 33.9 & 20.5 & 7.9 & 0.8 & 46.4 & 35.6 & 2.9 & 2.3 \\
  JSRNet\cite{Vojir2021road} & ICCV'21 & 33.6 & 43.9 & 20.2 & 29.3 & 13.7 & 28.1 & 28.9 & 18.6 & 24.5 & 11.0 \\
  Road Inpainting\cite{Lis2020detecting} & arXiv'20 & - & - & - & - & - & 54.1 & 47.1 & 57.6 & 39.5 & 36.0 \\
  Image Resyn.\cite{lis2019detecting} & ICCV'19 & 52.3 & 25.9 & 39.7 & 11.0 & 12.5 & 37.7 & 4.7 & 16.6 & 20.5 & 8.4 \\
  ObsNet\cite{Besnier2021triggering} & ICCV'21 & 75.4 & 26.7 & 44.2 & 52.6 & 45.1 & - & - & - & - & - \\
  NFlowJS\cite{Grcic2021dense} & arXiv'21 & 56.9 & 34.7 & 36.9 & 18.0 & 14.9 & 85.6 & 0.4 & 45.5 & 49.5 & 50.4 \\
  Max. Entropy \cite{Li2019expectation} & ICCV'19 & 85.5 & 15.0 & 49.2 & 39.5 & 28.7 & 85.1 & 0.8 & 47.9 & 62.6 & 48.5 \\
  DenseHybrid\cite{Grcic2022densehybrid} & ECCV'22 & 78.0 & 9.8 & 54.2 & 24.1 & 31.1 & 87.1 & 0.2 & 45.7 & 50.1 & 50.7 \\
  PEBAL\cite{Tian2022pixelwise} & ECCV'22 & 49.1 & 40.8 & 38.9 & 27.2 & 14.5 & 5.0 & 12.7 & 29.9 & 7.6 & 5.5 \\
  SynBoost\cite{DiBiase2021pixelwise} & CVPR'21 & 56.4 & 61.9 & 34.7 & 17.8 & 10.0 & 71.3 & 3.2 & 44.3 & 41.8 & 37.6 \\
  Mask2Anomaly\cite{rai2024mask2anomaly} & TPAMI'24 & \underline{88.7} & 14.6 & \underline{55.2} & 51.6 & 47.1 & 93.2 & 0.2 & 55.7 & 75.4 & 68.1 \\
  \hline
  RPL\cite{liu2023residual} & ICCV'23 & 83.4 & 11.7 & 49.7 & 29.9 & 30.1 & 85.9 & 0.6 & 52.6 & 56.6 & 56.6 \\
  +Ours & - & 87.3 & \underline{7.8} & 48.1 & \underline{52.4} & \textbf{56.1} & \underline{94.1} & \underline{0.06} & \underline{67.7} & \textbf{81.3} & \underline{86.5} \\
  \hline
  RbA\cite{nayal2023rba} & ICCV'23 & \textbf{90.9} & 11.6 & \textbf{55.7} & 52.1 & 46.8 & 91.8 & 0.5 & 58.4 & 58.8 & 60.9 \\
  +Ours & - & 85.6 & \textbf{7.7} & 46.2 & \textbf{55.2} & \underline{54.9} & \textbf{94.5} & \textbf{0.05} & \textbf{73.0} & \underline{80.4} & \textbf{89.9} \\
  \bottomrule
  \end{tabular}
  }
\end{table*}
\section{Experiments}
\subsection{Experiments Setup}
\paragraph{Baseline Methods.} 
The proposed framework demonstrates excellent transferability and seamlessly integrates with anomaly segmentation methods. To validate the performance improvement brought by the proposed framework to different anomaly segemntation methods, RPL  \cite{liu2023residual} and RbA  \cite{nayal2023rba} are adopted as baselines for comparative experiments.
RPL introduces a residual pattern learning module and employs a context-robust contrastive learning method to assign anomaly scores at the pixel level. RbA designs a novel anomaly scoring function that assigns anomaly scores at the pixel level by rejecting all known categories. Additionally, GroundingDINO\cite{liu2024grounding} is incorporated as an anomaly detection baseline.

\paragraph{Implementation Details.} 
In our framework, the anomaly segmentation models (RPL\cite{liu2023residual} and RbA\cite{nayal2023rba}) are frozen, while the remaining networks are jointly trained for 100 epochs using AdamW\cite{loshchilov2017decoupled} with a batch size of 16. The initial learning rate is set to \(1\times10^{-5}\), decaying every 10 epochs, and scheduled by StepLR\cite{gpy2017stepLR}. The loss weights \(\lambda_{\text{detect}}\), \(\lambda_{\text{orth}}\), and \(\lambda_{\text{ADT}}\) are set to 0.5, 0.1, and 0.1, respectively. \zj{The feature fusion module comprises 4 stacked layers, each equipped with 8 parallel attention heads and configured with a dropout rate of 0.1 for regularization.}

\paragraph{Evaluation Metrics.} 
In the Road Anomaly dataset, we conducted comprehensive evaluations using three complementary metrics to ensure thorough performance assessment: average precision (AP) capturing the overall detection accuracy across varying confidence thresholds, the area under the ROC curve (AuROC) measuring the model's discrimination ability irrespective of class imbalance, and the false positive rate at a 95\% true positive rate threshold (FPR95) quantifying false alarm rates when maintaining high detection sensitivity. For the SMIYC benchmark, we followed the established evaluation protocol that reports both pixel-level and component-level metrics \cite{chan2021segmentmeifyoucan}. Please refer to Appendix~\ref{sec:sec2} for the evaluation protocol of component-level metrics on the SMIYC dataset.

\paragraph{Datasets.} 
Following S2M\cite{zhao2024segment}, the synthetic dataset is employed to train the proposed framework. In addition, three datasets are used to validate the effectiveness of the proposed method. 
\textbf{Segment Me If You Can (SMIYC)} \cite{chan2021segmentmeifyoucan} includes two subsets, the AnomalyTrack and the ObstacleTrack. The AnomalyTrack contains 100 images of unknown objects of various sizes in different environments. The ObstacleTrack consists of 412 images, typically depicting small unknown objects on roads, 85 of which are captured under nighttime or adverse weather conditions. SMIYC is a publicly available benchmark whose leaderboard can be viewed on a public webpage.
\textbf{RoadAnomaly} \cite{lis2019detecting} contains 60 real-world images from various online platforms. These images depict anomalous objects near vehicles, such as wildlife, debris, abandoned tires, waste containers, and construction machinery. Each image is meticulously annotated at the pixel level to identify the precise locations of the anomalous objects.
\begin{table}[t]
\centering
\caption{\textbf{Comparison with state-of-the-art methods on RoadAnomaly dataset.}}
\label{tab:road_anomaly}
\renewcommand{\arraystretch}{1.15}
\resizebox{\columnwidth}{!}{
\begin{tabular}{l|c|ccc}
\toprule
\textbf{Method} & \textbf{Venue} & \textbf{FPR95 $\downarrow$} & \textbf{AP $\uparrow$} & \textbf{AuROC $\uparrow$} \\
\midrule
Max softmax\cite{hendrycks2017a} & ICLR'17 & 68.15 & 22.38 & 75.12 \\
Gambler\cite{Liu2019deepgamblers} & NeurIPS'19 & 48.79 & 31.45 & 85.45 \\
SynthCP \cite{Xia2020synthesize}& ECCV'20 & 64.69 & 24.86 & 76.08 \\
Synboost \cite{DiBiase2021pixelwise}& ICCV'21 & 59.72 & 41.83 & 85.23 \\
SML\cite{Jung2021standardized} & CVPR'21 & 49.74 & 25.82 & 81.96 \\
GMMSeg \cite{Liang2022gmmseg} & NeurIPS'22 & 47.90 & 34.42 & 84.71 \\
PEBAL\cite{Tian2022pixelwise} & ECCV'22 & 44.58 & 45.10 & 87.63 \\
MGCDA\cite{zhang2023improving} & MM'23 & 42.19 & 50.35 & - \\
\hline
RPL \cite{liu2023residual}& ICCV'23 & 17.74 & 71.60 & 95.72 \\
+Ours & - & \textbf{4.78} & \textbf{87.13} & \textbf{98.73} \\
\hline
RbA\cite{nayal2023rba}& ICCV'23 & 6.92 & 85.42 & 97.99 \\ 
+Ours & - & \textbf{2.11} & \textbf{95.21} & \textbf{98.94} \\
\bottomrule
\end{tabular}
}
\end{table}
\subsection{Comparisons with SOTA methods}
\subsubsection{Segment Me If You Can Benchmark}
As shown in Table~\ref{tab:smlyc_results}, the most notable achievement of our method is the significant reduction in the FPR95. 
Specifically, the model integrated with RPL achieves 7.8\% FPR95 on AnomalyTrack and 0.06\% FPR95 on ObstleTrack, showing a significant improvement over the baseline RPL (11.7\% and 0.6\%, respectively). 
Similarly, the model integrated with RbA reduces FPR95 to 7.7\% and 0.05\% on their respective datasets, demonstrating consistent improvements over the original RbA implementation (11.6\% and 0.5\%). The two baseline methods directly predict per-pixel anomaly score maps to identify anomalous pixels. However, this approach is prone to background noise, leading to false positive results. Our framework aims to mitigate these false positives by transitioning from object-level anomaly detection to pixel-level fine-grained anomaly selection. These results validate that our proposed framework effectively reduces false positives in real-world test datasets across various scenarios.

Meanwhile, we achieve state-of-the-art (SOTA) results on most metrics\footnote{https://segmentmeifyoucan.com/leaderboard}, with the AP on AnomalyTrack increasing from 83.4\% to 87.3\%, and the AP on ObstleTrack improving from 85.9\% to 94.1\%. While maintaining competitive AP scores, we also significantly improve component-level metrics, especially PPV and average F1 scores.  These improvements stem from our coarse-to-fine anomaly segmentation paradigm. The OUAFS module enhances spatial coherence through orthogonal uncertainty fusion, while ADT-Net optimizes region-specific thresholds. Together, they preserve complete anomaly structures while effectively eliminating false positives.
\subsubsection{RoadAnomaly Benchmark}
As shown in Table~\ref{tab:road_anomaly}, the model integrated with RPL reduces FPR95 from 17.74\% to 4.78\%. The model integrated with RbA achieves an impressive FPR95 of 2.11\%. These results show substantial improvements compared to traditional methods, such as PEBAL \cite{Tian2022pixelwise} (44.58\%) and GMMSeg \cite{Liang2022gmmseg} (47.90\%). Furthermore, our framework also substantially improves the AP metric, increasing from 71.6\% to 87.1\% for RPL (15.5\% improvement) and from 85.4\% to 95.2\% for RbA (9.8\% improvement). We achieve state-of-the-art results across all evaluation metrics\footnote{ https://paperswithcode.com/sota/anomaly-detection-on-road-anomaly}, with our best configuration reaching 98.9\% AuROC.

\begin{figure*}
  \includegraphics[width=1\textwidth]{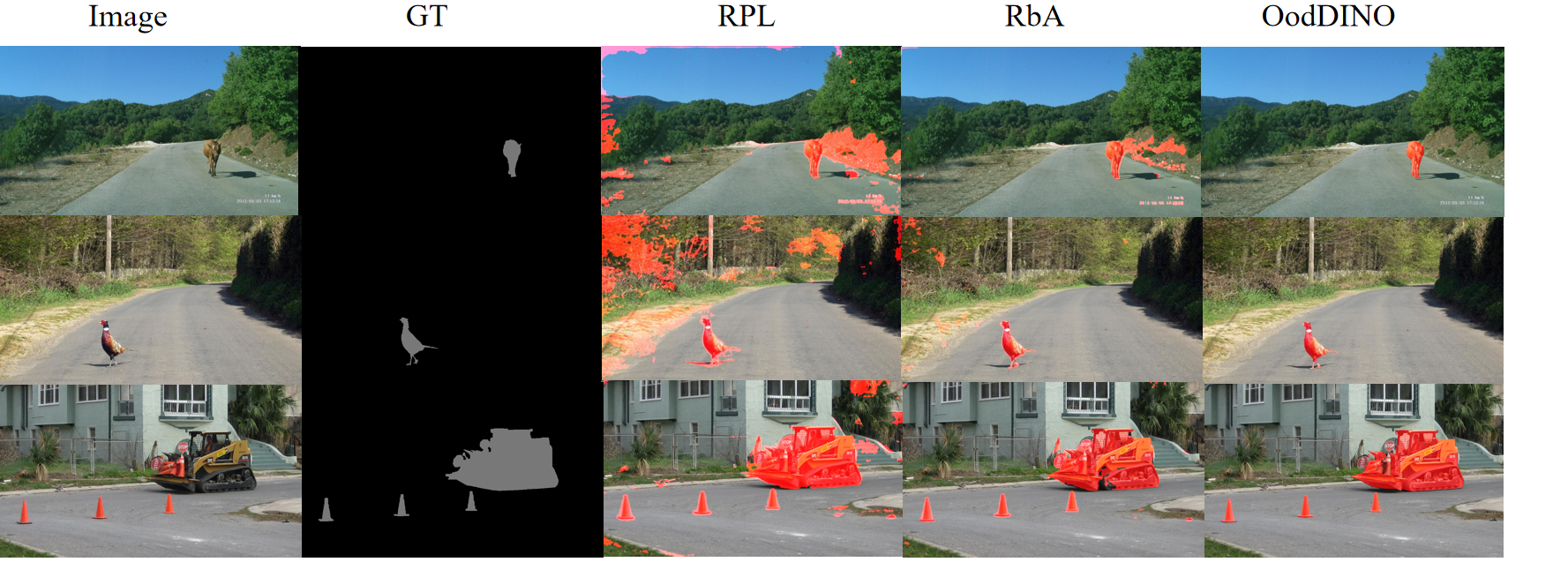} 
  \caption{\textbf{Qualitative comparison of anomaly segmentation methods.}The two leftmost columns display the image and its
ground truth. Additionally, the anomaly predictions from RPL\cite{liu2023residual} and RbA\cite{nayal2023rba} (third and fourth columns), as well as our method (last
column), highlight the high-score anomaly map (in red), indicating anomalous pixels.}
  \label{fig:fig1} 
\end{figure*}

Overall, across 13 metrics on the three datasets, the framework integrated with RPL achieved improvements in 12 metrics, while the method integrated with RbA demonstrated improvements in 11 metrics. Notably, the FPR95 for both baselines has been significantly reduced across all three datasets. As illustrated in Fig.~\ref{fig:fig1}, the proposed framework demonstrates significant improvements over both baseline methods by substantially reducing noise in prediction results while accurately localizing pixel-level anomalous regions. These quantitative and qualitative experimental results collectively validate the effectiveness of incorporating object spatial priors and region-specific thresholds in addressing both background noise interference and object fragmentation issues. Additional qualitative results are provided in Appendix~\ref{sec:6}.

\subsection{Ablation Studies\label{sec:abl}}
In this section, ablation studies are conducted on the SMIYC (AnomalyTrack and ObstacleTrack) and RoadAnomaly datasets to demonstrate the effectiveness of the proposed modules in Sec. \ref{sec:method}. As mentioned above, the RPL\cite{liu2023residual} are adopted as the baselines for our experiments.
\begin{table}[htbp]
\caption{\textbf{Performance comparison on three benchmarks (RA: RoadAnomaly, AT: AnomalyTrack, OT: ObstacleTrack). GD denotes Grounding DINO. FPR denotes FPR95.}}
\centering
\setlength{\tabcolsep}{3pt}
\renewcommand{\arraystretch}{1.1}
\resizebox{\columnwidth}{!}{
\begin{tabular}{l|cc|cc|cc}
\hline
\multirow{2}{*}{\makecell{Method}} & \multicolumn{2}{c|}{RA} & \multicolumn{2}{c|}{AT} & \multicolumn{2}{c}{OT} \\
\cline{2-7}
 & FPR↓ & AP↑ & FPR↓ & AP↑ & FPR↓ & AP↑ \\
\hline
RPL & 17.74 & 71.60 & 7.18 & 88.55 & 0.09 & 96.91 \\
RPL + GD & 28.50 & 63.33 & 21.25 & 75.47 & 15.30 & 77.12 \\
RPL + GD + ADT-Net & 12.58 & 76.60 & 6.50 & 89.05 & 0.08 & 97.10\\
RPL + GD + OUAFS & 9.58 & 85.11 & 6.73 & 88.92 & 0.07& 97.14 \\
OoDDINO & \textbf{4.78} & \textbf{87.13} & \textbf{3.82} & \textbf{92.08} & \textbf{0.05} & \textbf{97.71} \\
\hline
\end{tabular}
}
\label{tab:Table3}
\end{table}

\textbf{Effects of Framework Components:}
As shown in Table~\ref{tab:Table3}, we compare five network configurations to evaluate each component's contribution. The configurations include: (a) Baseline: original RPL implementation; (b) RPL+GD: integrating Grounding DINO \cite{liu2024grounding} to predict anomaly bounding boxes, with pixels outside boxes classified as normal; (c) RPL+GD+ADT: incorporating our Adaptive Dual-Threshold Network to dynamically generate region-specific thresholds; (d) RPL+GD+OUAFS: integrating our Orthogonal Uncertainty-Aware Fusion Strategy with Grounding DINO to enhance the model's anomaly detection performance; and (e) OoDDINO: our complete framework combining all components.

Compared to RPL, the performance of RPL+GD decreases across all metrics. We attribute this decline to two primary factors: (1) the limited capability of Grounding DINO in detecting anomalous objects, leading to missed or false detections. (2) the direct use of erroneous results from Grounding DINO to guide the distinction between normal and anomalous regions, negatively impacting anomaly segmentation outcomes. By introducing ADT-Net, RPL+GD+ADT shows significant improvements, with FPR95 decreasing by 15.1\% and AP increasing by 13.9\% on AnomalyTrack. This demonstrates that ADT-Net enhances the framework's fault tolerance by adaptively integrating object-level detection results and pixel-level segmentation results, enabling anomalous regions to be finely classified. Furthermore, the integration of OUAFS leads to substantial performance improvements for RPL+GD+OUAFS compared to RPL+GD. This result indicates that OUAFS enhances Grounding DINO's ability to detect anomalous objects by incorporating uncertainty information, enabling RPL to more accurately distinguish between normal and anomalous regions. The complete OoDDINO framework achieves optimal results across all metrics on the three datasets, significantly outperforming the baseline. These results validate our coarse-to-fine anomaly segmentation paradigm that progressively refines anomaly localization from object-level detection to pixel-wise selection.

\begin{table}[h!]
\centering
\caption{\textbf{Comparative Analysis of Different Fusion Strategies on Anomaly Detection Performance.}}
\vspace{-3mm}
\label{tab:afem_ablation}
\resizebox{1\columnwidth}{!}{
\begin{tabular}{@{}lccccl@{}}
\toprule
\multirow{2}{*}{\textbf{Dataset}} & \multicolumn{3}{c}{\textbf{mAP}} & \multirow{2}{*}{\textbf{Fusion Strategy}} \\
\cmidrule(lr){2-4}
& \textbf{Small} & \textbf{Medium} & \textbf{Large} & \\ \midrule
\multirow{5}{*}{RoadAnomaly} 
& 0.435 & 0.815 & 0.820 & Img + Seg \\
& 0.440 & 0.820 & 0.830 & Img + Entropy \\
& 0.445 & 0.830 & 0.840 & Img + Dis \\
& 0.480 & 0.870 & 0.885 & Img+Seg+Entropy+Dis \\
& \textbf{0.495} & \textbf{0.880} & \textbf{0.900} & Ours \\ \midrule
\multirow{5}{*}{ObstacleTrack}
& 0.420 & 0.800 & 0.810 & Img + Seg \\
& 0.425 & 0.810 & 0.820 & Img + Entropy \\
& 0.430 & 0.815 & 0.830 & Img + Dis \\
& 0.450 & 0.840 & 0.855 & Img+Seg+Entropy+Dis \\
& \textbf{0.465} & \textbf{0.850} & \textbf{0.870} & Ours \\ \midrule
\multirow{5}{*}{AnomalyTrack}
& 0.410 & 0.790 & 0.800 & Img + Seg \\
& 0.415 & 0.800 & 0.810 & Img + Entropy \\
& 0.420 & 0.805 & 0.820 & Img + Dis \\
& 0.440 & 0.830 & 0.845 & Img+Seg+Entropy+Dis \\
& \textbf{0.455} & \textbf{0.850} & \textbf{0.865} & Ours \\ \bottomrule
\end{tabular}
}
\end{table}

\textbf{Effects of Fusion Strategies:}
To evaluate the effectiveness of different feature fusion strategies in the proposed OUAFS, we conduct experiments on three datasets to analyze the model's performance in detecting OoD objects of different scales.

As shown in Table~\ref{tab:afem_ablation}, among the single-modality fusion strategies, the model achieves the most significant performance improvement by incorporating distance map features, followed by entropy maps and semantic segmentation maps. This indicates that confidence-based distance information provides the most complementary features for image representation. When the three uncertainty maps are integrated into the model in parallel, a significant performance boost is observed. For instance, on the RoadAnomaly dataset, this full fusion strategy achieves mAP scores of 0.480/0.870/0.885, representing an improvement of approximately 0.035–0.045 compared to single-modality fusion methods.

By introducing the three uncertainty maps sequentially, OUAFS further enhances the model’s detection performance, achieving the best results across all datasets. On the RoadAnomaly dataset, OUAFS achieves an mAP score of 0.495/0.880/0.900, which is approximately 0.015 higher than the full fusion strategy. Similar improvements are observed on the ObstacleTrack and AnomalyTrack datasets, where our approach consistently outperforms all baseline methods. These experimental results demonstrate: (1) uncertainty information positively contributes to enhancing the ability of open-set object detection models to detect anomalous objects. (2) compared to parallel fusion strategies, sequential feature fusion effectively complements anomalous information in images, leading to superior detection performance.

\begin{table}[htbp]
\caption{\textbf{Performance comparison of different normalization strategies across three benchmarks. FPR denotes FPR95.}}
\vspace{-3mm}
\centering
\setlength{\tabcolsep}{3pt}
\renewcommand{\arraystretch}{1.1}
\resizebox{\columnwidth}{!}{
\begin{tabular}{l|cc|cc|cc}
\hline
\multirow{2}{*}{Normalization Strategy} & \multicolumn{2}{c|}{RoadAnomaly} & \multicolumn{2}{c|}{AnomalyTrack} & \multicolumn{2}{c}{ObstacleTrack} \\
\cline{2-7}
 & FPR↓ & AP↑ & FPR↓ & AP↑ & FPR↓ & AP↑ \\
\hline
No Normalization & 18.43 & 72.35 & 9.72 & 84.68 & 0.21 & 91.45 \\
Global Linear & 15.64 & 75.92 & 8.15 & 87.23 & 0.17 & 93.61 \\
Global Sigmoid & 11.27 & 79.30 & 7.21 & 88.75 & 0.14 & 95.35 \\
Region-Separate & 8.95 & 82.64 & 5.78 & 90.12 & 0.11 & 96.28 \\
Region-Adaptive (Ours) & \textbf{4.78} & \textbf{87.13} & \textbf{3.82} & \textbf{92.08} & \textbf{0.05} & \textbf{97.71} \\
\hline
\end{tabular}
}
\label{tab:normalization}
\end{table}
\textbf{Effects of Anomaly Score Normalization Strategies:}
We evaluate five normalization strategies in our ADT-Net, as shown in Table~\ref{tab:normalization}. By employing normalization methods (global linear or global sigmoid), the model's performance is enhanced, demonstrating that normalizing anomaly scores to specific ranges facilitates more accurate threshold prediction. The region-separate strategy, which normalizes foreground and background regions independently, yields further performance gains. These results validate our key observation that anomaly score distributions vary significantly across different regions. Our region-adaptive approach, which dynamically learns optimal parameters for different regions, consistently outperforms all alternatives. These improvements demonstrate that region-adaptive normalization effectively handles score distribution discrepancies between different methods, enabling our framework to learn more accurate thresholds while maintaining compatibility with various baseline architectures.

\section{Conclusion}
In this paper, we presented OoDDINO, a novel multi-level anomaly segmentation framework that addresses the limitations of existing pixel-wise approaches through a coarse-to-fine detection strategy. By integrating an uncertainty-guided object-level detector and a pixel-level segmentation model in a two-stage cascade architecture, OoDDINO effectively captures both spatial priors and fine-grained details of anomalous regions. Our proposed Orthogonal Uncertainty-Aware Fusion Strategy (OUAFS) enhances the localization capability of the detection stage, while the Adaptive Dual-Threshold Network (ADT-Net) enables region-aware segmentation with dynamic thresholding for foreground and background areas. Notably, OoDDINO is compatible with various pixel-wise anomaly detection methods and can serve as a plug-in module to enhance their performance. Extensive experiments on public benchmarks demonstrate that OoDDINO achieves state-of-the-art results, highlighting its robustness, adaptability, and practical value for real-world anomaly segmentation tasks.

\begin{acks}
This work was supported by the Research and Practice of Software Engineering, a key technology project for intelligent rural development on the Qinghai-Tibet Plateau (Grant No. 2024CXTD09), and by the Sichuan Science and Technology Program (Grant No. 2023YFN0026).
\end{acks}

\bibliographystyle{ACM-Reference-Format}
\bibliography{sample-base}

\appendix
\section*{Appendix}

In this appendix, we provide additional experimental results, evaluation details, and qualitative visualizations to support the findings in the main paper. Specifically, Section~\ref{sec:1} reports detailed results on both subsets of the Fishyscapes~\cite{blum2021fishyscapes} validation set. Section~\ref{sec:sec2} introduces the component-level evaluation metrics used for the SMIYC~\cite{chan2021segmentmeifyoucan} dataset. Section~\ref{sec:sec3} presents an ablation study on the effect of normalized score ranges. Section~\ref{sec:sec4} demonstrates the transferability of our method on more baseline models. Section~\ref{sec:sec5} analyzes the computational efficiency of our framework. Finally, Section~\ref{sec:6} provides qualitative visualizations on multiple datasets.

\section{Experiments on Fishyscapes \label{sec:1}}

\begin{table*}[htbp]
\centering
\caption{Comparison with state-of-the-art methods on the Fishyscapes benchmark. Best results are in \textbf{bold}.}
\begin{tabular}{l|c|ccc|ccc}
\toprule
\multirow{2}{*}{\textbf{Methods}} & \multirow{2}{*}{\textbf{Venue}} & \multicolumn{3}{c|}{\textbf{Static}} & \multicolumn{3}{c}{\textbf{Lost \& Found}} \\
 & & \textbf{FPR ↓} & \textbf{AP ↑} & \textbf{AUROC ↑} & \textbf{FPR ↓} & \textbf{AP ↑} & \textbf{AUROC ↑} \\
\midrule
Maximum Softmax~\cite{hendrycks2017a} & ICLR'17 & 23.31 & 26.77 & 93.14 & 10.36 & 40.34 & 90.82 \\
Mahalanobis~\cite{lee2018simple} & NeurIPS'18 & 11.70 & 27.37 & 96.76 & 11.24 & 56.57 & 96.75 \\
SML~\cite{Jung2021standardized} & CVPR'21 & 12.14 & 66.72 & 97.25 & 33.49 & 22.74 & 94.97 \\
SynBoost~\cite{di2021pixel} & ICCV'21 & 25.59 & 66.44 & 95.87 & 31.02 & 60.58 & 96.21 \\
Meta-OoD~\cite{chan2021entropy} & ICCV'21 & 13.57 & 72.91 & 97.56 & 37.69 & 41.31 & 93.06 \\
DenseHybrid~\cite{Grcic2022densehybrid} & ECCV'22 & 4.17 & 76.23 & 99.07 & 5.09 & 69.79 & 99.01 \\
PEBAL~\cite{Tian2022pixelwise} & ECCV'22 & 1.52 & 92.08 & 99.61 & 4.76 & 58.81 & 98.96 \\
RPL~\cite{liu2023residual} (Baseline) & ICCV'23 & 0.85 & 92.46 & 99.73 & 2.52 & 70.61 & 99.39 \\
\hline
\textbf{Ours} & - & \textbf{0.27} & \textbf{95.26} & \textbf{99.80} & \textbf{0.06} & \textbf{93.12} & \textbf{99.42} \\
\bottomrule
\end{tabular}
\label{tab:fishyscapes_results}
\end{table*}

The Fishyscapes~\cite{blum2021fishyscapes} dataset is a standard benchmark designed to evaluate the capability of semantic segmentation models to detect anomalous objects in open-world environments. The \textit{Static} subset introduces out-of-distribution (OoD) objects into urban street scenes, while the \textit{Lost \& Found} subset focuses on small, sparsely distributed anomalous objects in road environments.

As shown in Table~\ref{tab:fishyscapes_results}, our method significantly outperforms prior approaches on both subsets. On the \textit{Static} subset, our method reduces the FPR from 0.85 (RPL baseline) to 0.27 and improves AP from 92.46 to 95.26. On the \textit{Lost \& Found} subset, we reduce the FPR from 2.52 to 0.06 and achieve an AP of 93.12. These results demonstrate the effectiveness of our coarse-to-fine anomaly classification strategy in reducing false positives and improving detection accuracy.

\section{Evaluation Metrics \label{sec:sec2}}

In addition to pixel-level metrics such as False Positive Rate (FPR) and Average Precision (AP), we adopt three component-level metrics to better assess anomaly detection performance on the SMIYC dataset~\cite{chan2021segmentmeifyoucan}.

We define component-wise true positives (TP), false negatives (FN), and false positives (FP), based on an adjusted version of the component-wise intersection over union (sIoU)~\cite{blum2021fishyscapes}. For each ground-truth component \(k\), the sIoU is computed as:
\begin{equation}
\text{sIoU}(k) := \frac{|k \cap \hat{K}(k)|}{|(k \cup \hat{K}(k)) \setminus A(k)|},
\end{equation}
where \(\hat{K}(k)\) is the union of predicted components overlapping with \(k\), and \(A(k)\) excludes pixels belonging to other ground-truth components.

A component is counted as a TP if \(\text{sIoU} > \tau\), and as an FN otherwise. For predicted components, we define precision (PPV) as:
\begin{equation}
\text{PPV}(\hat{k}) := \frac{|\hat{k} \cap K(\hat{k})|}{|\hat{k}|},
\end{equation}
and classify \(\hat{k}\) as FP if \(\text{PPV} \leq \tau\).

The final component-level F1-score is given by:
\begin{equation}
\text{F1}(\tau) := \frac{2 \cdot \text{TP}}{2 \cdot \text{TP} + \text{FN} + \text{FP}},
\end{equation}
which balances detection accuracy and localization quality.

\section{Effect of Normalized Interval on the Framework \label{sec:sec3}}

In Section~\ref{sec:ADTGN}, we set the parameter \(\alpha\) in Equation~\ref{eq:eq} empirically to 0.3. Specifically, we fix the normalized score range length to 0.5 and vary the lower bound from 0.1 to 0.4 in steps of 0.1. As shown in Table~\ref{tab:score_range_ablation}, the score range \([0.3, 0.8]\) yields the best results, achieving the lowest FPR of 4.78\% and the highest AP of 87.13\%.

\begin{table}[htbp]
\centering
\caption{Performance comparison under different score range settings. Best results in \textbf{bold}.}
\begin{tabular}{l|c|c}
\toprule
\textbf{Score Range} & \textbf{FPR ↓} & \textbf{AP ↑} \\
\midrule
$[0.1, 0.6]$ & 6.45 & 81.73 \\
$[0.2, 0.7]$ & 5.98 & 84.20 \\
$[0.4, 0.9]$ & 5.57 & 84.01 \\
$[0.3, 0.8]$ & \textbf{4.78} & \textbf{87.13} \\
\bottomrule
\end{tabular}
\label{tab:score_range_ablation}
\end{table}

\section{Experiments on More Baselines \label{sec:sec4}}

To demonstrate the transferability of our framework, we integrate it into two representative baselines: PEBAL~\cite{Tian2022pixelwise} and Mask2Anomaly~\cite{rai2024mask2anomaly}, and evaluate on the RoadAnomaly test set. As shown in Table~\ref{tab:roadanomaly_results}, our method improves the AP by 34.6\% and 5.96\%, and reduces the FPR by 23.18\% and 6.12\%, respectively. These results validate the generalizability and robustness of our approach.

\begin{table}[htbp]
\centering
\caption{Performance comparison on RoadAnomaly test set. Best results in \textbf{bold}.}
\begin{tabular}{l|c|c}
\toprule
\textbf{Method} & \textbf{AP ↑} & \textbf{FPR ↓} \\
\midrule
PEBAL~\cite{Tian2022pixelwise} & 45.10 & 44.58 \\
PEBAL + Ours & 71.11 & 21.40 \\
\hline
Mask2Anomaly~\cite{rai2024mask2anomaly} & 79.70 & 13.45 \\
Mask2Anomaly + Ours & \textbf{85.66} & \textbf{7.33} \\
\bottomrule
\end{tabular}
\label{tab:roadanomaly_results}
\end{table}

\section{Efficiency Analysis \label{sec:sec5}}

As shown in Table~\ref{tab:efficiency_comparison}, our method introduces additional computational overhead due to the enhanced architecture. With 660M parameters and 410.88 GFLOPs, our model is larger than RPL (168M parameters, 32.1 GFLOPs). However, the inference speed only decreases from 4.56 FPS to 3.51 FPS on the SMIYC dataset. The proposed method can still meet real-time requirements for autonomous driving when integrated with frame selection or keyframe strategies.

\begin{table}[htbp]
\centering
\caption{Comparison of computational cost and inference speed with the RPL~\cite{liu2023residual} baseline.}
\begin{tabular}{lccc}
\toprule
\textbf{Method} & \textbf{\#Params (M)} & \textbf{GFLOPs} & \textbf{FPS} \\
\midrule
RPL (Baseline) & 168 & 32.10 & 4.56 \\
Ours           & 660 & 410.88 & 3.51 \\
\bottomrule
\end{tabular}
\label{tab:efficiency_comparison}
\end{table}

\section{Qualitative Results \label{sec:6}}

\begin{figure*}[htbp]
\centering
\includegraphics[scale=0.81]{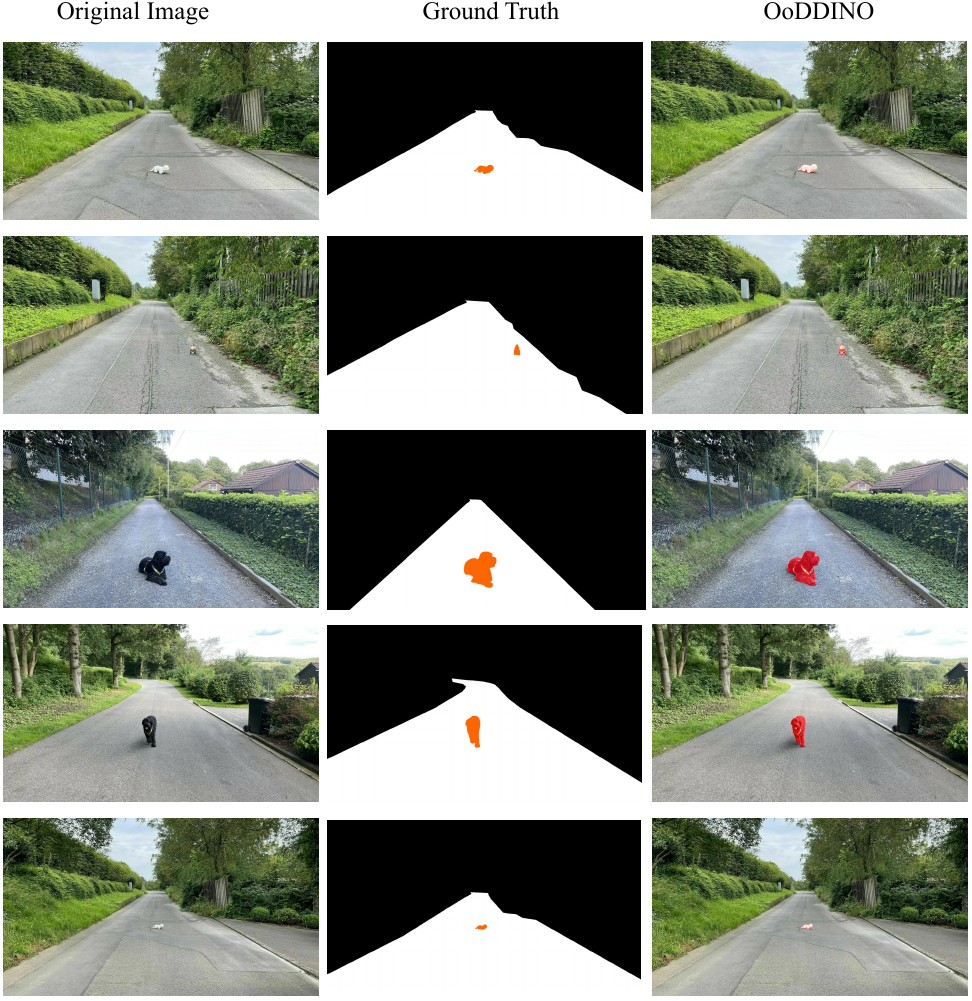}
\vspace{-0.1in}
\caption{Qualitative results on the ObstacleTrack dataset. Left: original images. Middle: ground-truth annotations. Right: predicted anomaly segmentation maps.}
\label{fig:trancos_vis}
\vspace{-0.1in}
\end{figure*}

\begin{figure*}[htbp]
\centering
\includegraphics[scale=0.81]{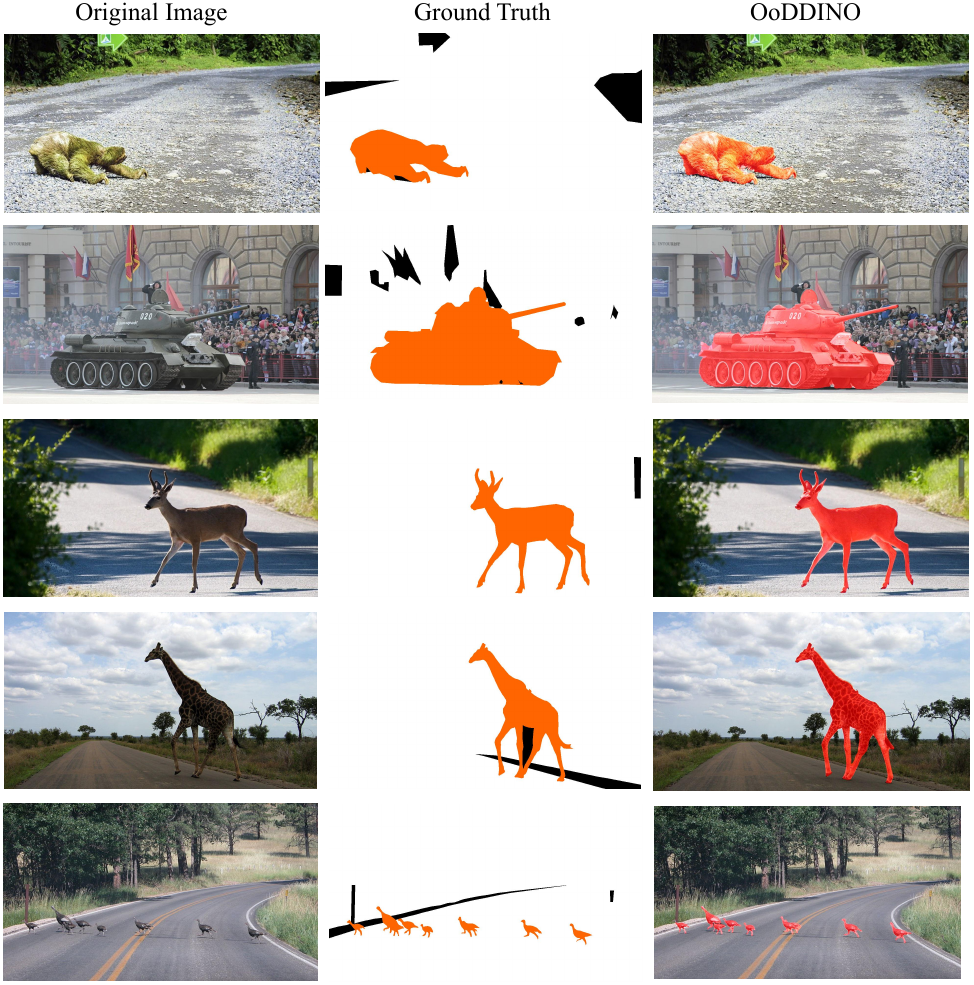}
\vspace{-0.1in}
\caption{Qualitative results on the AnomalyTrack dataset. Left: original images. Middle: ground-truth annotations. Right: predicted anomaly segmentation maps.}
\label{fig:vis}
\vspace{-0.1in}
\end{figure*}

We provide additional visual results generated by OoDDINO on SMIYC~\cite{chan2021segmentmeifyoucan} datasets to illustrate the high-quality anomaly segmentation achieved by our model. Predicted segmentation maps are shown for both the ObstacleTrack and AnomalyTrack subsets in Figures~\ref{fig:trancos_vis} and~\ref{fig:vis}, respectively.

\end{document}